\documentclass[12pt]{article}
\usepackage{amsfonts}
\usepackage{a4wide}
\usepackage{graphicx}
\usepackage{amsmath,amscd,amsthm,a4,amssymb}

\begin{document}

\begin{center}
{\Large \textbf{Approximation error of single hidden layer neural networks
with fixed weights}}

\

\textbf{Vugar E. Ismailov} \

{Institute of Mathematics and Mechanics}

Azerbaijan National Academy of Sciences

Az-1141, Baku, Azerbaijan

{e-mail:} {vugaris@mail.ru}
\end{center}

\smallskip

\textbf{Abstract.} \textit{This paper provides an explicit formula for the
approximation error of single hidden layer neural networks with two fixed
weights.}

\smallskip


\smallskip


\

\

\centerline{\textbf{1. INTRODUCTION}}

\

Properties of feedforward neural networks with one hidden layer have been
studied quite well. By selecting different activation functions, many
authors showed that single hidden layer neural networks possess the
universal approximation property. In recent years, the theory of neural
networks has been developed further in this direction. For example, from the
point of view of practical applications, neural networks with a restricted
set of weights have gained special interest.

A \textit{single hidden layer neural network} with $r$ units in the hidden
layer and input $\mathbf{x}=(x_{1},...,x_{d})$ computes a function of the
form
\begin{equation*}
\sum_{i=1}^{r}c_{i}\sigma (\mathbf{w}^{i}\mathbf{\cdot x}-\theta _{i}),\eqno%
(1.1)
\end{equation*}%
where the \textit{weights} $\mathbf{w}^{i}$ are vectors in $\mathbb{R}^{d}$,
the \textit{thresholds} $\theta _{i}$ and the \textit{coefficients} $c_{i}$
are real numbers and \ the \textit{activation function} $\sigma $ is a real
univariate function. For various activation functions $\sigma $, it was
shown by many authors that one can approximate arbitrarily well to any
continuous function by functions of the form (1.1) ($r$ is not fixed!) over
any compact subset of $\mathbb{R}^{d}$. That is, the set
\begin{equation*}
\mathcal{M}(\sigma )=span\text{\ }\{\sigma (\mathbf{w\cdot x}-\theta ):\
\theta \in \mathbb{R}\text{, }\mathbf{w}\in \mathbb{R}^{d}\}
\end{equation*}%
is dense in the space $C(\mathbb{R}^{d})$ in the topology of uniform
convergence on compact sets (see, e.g., \cite{A2,A3,A4,A7,A10,A20}). The
most general and complete result of this type was obtained by Leshno, Lin,
Pinkus and Schocken \cite{A13}. They proved that a continuous activation
function has the \textit{density property} or the \textit{universal
approximation property} if and only if it is not a polynomial. This result
has shown the power of single hidden layer neural networks within all
possible choices of the continuous activation function $\sigma $. For
detailed information on this and other density results see \cite{A20}.

It was formerly believed and particularly emphasized in many works that for
the universal approximation property, large networks with sufficiently many
hidden neurons are needed. However, the recent papers \cite{GI2,GI3} have
shown that there exist neural networks with very few hidden neurons, which
can approximate arbitrarily well any continuous function on any compact set.
Moreover, it was shown that such networks can be constructed in practice.

A number of authors proved that single hidden layer neural networks with
some suitably restricted set of weights also possess the universal
approximation property. For example, White and Stinchcombe \cite{A25} showed
that a single layer network with a polygonal, polynomial spline or analytic
activation function and a bounded set of weights has the universal
approximation property. Ito \cite{A10} investigated this property of
networks using a monotone \textit{sigmoidal function} (any continuous
function tending to $0$ at minus infinity and $1$ at infinity), with weights
located only on the unit sphere. Note that sigmoidal functions play an
important role in neural network theory and related application areas (see,
e.g., \cite{Cos1,Cos2,GI2,GI3,Il,Ky,Lin2,MP}). Thus we see that the weights
required for the universal approximation property are not necessarily of an
arbitrarily large magnitude. But what if they are too restricted. Obviously,
in this case, the universal approximation property does not hold, and the
problem reduces to the identification of compact subsets in $\mathbb{R}^{d}$
over which the model preserves its general propensity to approximate
arbitrarily well. The first and most interesting case is, of course, neural
networks with a finite set of weights. In \cite{Ismailov1}, we considered
this problem and gave sufficient and necessary conditions for good
approximation by networks with finitely many weights and also with weights
varying on finitely many straight lines. For a set $W$ of weights consisting
of two vectors or two straight lines, we showed that there is a
geometrically explicit solution to this problem (see \cite{Ismailov1}).

It should be remarked that the above density results do not tell about the
degree of approximation. They only provide us with the knowledge if and when
single hidden layer neural networks can approximate multivariate functions.
The problem of degree of approximation is related to the \textit{problem of
complexity}, which is the same as the problem of determining the number of
hidden neurons required for approximation within a given accuracy. This
problem was investigated in a number of papers (see, e.g., \cite%
{Barron,Hahm,Lin1,Lin2,Maiorov,Mhaskar}).

In this paper, we consider the uniform approximation of single hidden layer
networks with two fixed weights in $\mathbb{R}^{d}$. As noted above these
networks are not always dense in the space of continuous functions. In fact,
the possibility of density depends on a compact set, where all given
functions are defined. Characterization of compact sets, for which various
density results hold, was given in \cite{Ismailov1,IS}. Here we are
interested in the approximation error, the minimal number within which the
considered network can approximate a given multivariate function. We
establish an explicit approximation error formula for single hidden layer
neural networks with two fixed weights. Our formula is valid for many
activation functions. For example, it is valid for all continuous
nonconstant activation functions, which have limits at plus and minus
infinities.

\

\

\centerline{\textbf{2. THE MAIN RESULT}}

\

Assume $\sigma $ is a continuous function on $\mathbb{R}$. Assume, besides, $%
\mathbf{a}$ and $\mathbf{b}$ are two fixed nonzero vectors in $\mathbb{R}%
^{d} $. Consider the set

\begin{equation*}
\mathcal{N}(\sigma )=\left\{ \sum_{i=1}^{r}c_{i}\sigma (\mathbf{w}^{i}%
\mathbf{\cdot x}-\theta _{i}):r\in \mathbb{N},\text{ }c_{i},\theta _{i}\in
\mathbb{R}\right\},
\end{equation*}%
where the weights $\mathbf{w}^{i}=$ $\mathbf{a}$ or $\mathbf{w}^{i}=$ $%
\mathbf{b}$. That is, we consider the set of single hidden layer neural
networks with weights restricted to only two vectors. In particular, these
vectors may coincide, and then we have the set of neural networks with a
single fixed weight. Let $Q$ be a compact subset of $\mathbb{R}^{d}$ and $%
f\in C(Q).$ Consider the approximation of $f$ by neural networks from $%
\mathcal{N}(\sigma )$. The approximation error is defined as

\begin{equation*}
E\left( f,\mathcal{N}(\sigma )\right) \overset{def}{=}\inf_{\Lambda \in
\mathcal{N}(\sigma )}\left\Vert f-\Lambda \right\Vert .
\end{equation*}

The following objects, called \textit{paths}, were exploited in many papers.
We will use these objects in the further analysis.

\bigskip

\textbf{Definition 2.1.}\ \textit{A finite or infinite ordered set $\left(
\mathbf{p}{_{1},\mathbf{p}_{2},...}\right) \subset Q$ with $\mathbf{p}%
_{i}\neq \mathbf{p}_{i+1}$ and either $\mathbf{a}\cdot \mathbf{p}_{1}=%
\mathbf{a}\cdot \mathbf{p}_{2},\mathbf{b}\cdot \mathbf{p}_{2}=\mathbf{b}%
\cdot \mathbf{p}_{3},\mathbf{a}\cdot \mathbf{p}_{3}=\mathbf{a}\cdot \mathbf{p%
}_{4},...$ or $\mathbf{b}\cdot \mathbf{p}_{1}=\mathbf{b}\cdot \mathbf{p}%
_{2},~\mathbf{a}\cdot \mathbf{p}_{2}=\mathbf{a}\cdot \mathbf{p}_{3},\mathbf{b%
}\cdot \mathbf{p}_{3}=\mathbf{b}\cdot \mathbf{p}_{4},...,$ is called a path
with respect to the directions $\mathbf{a}$ and $\mathbf{b}$.}

\bigskip

It should be remarked that paths with respect to two directions in $\mathbb{R%
}^{2}$ were first considered by Braess and Pinkus \cite{3}. They proved a
theorem, which yields that the idea of paths are essential for deciding if a
set of points ${\left\{ {\mathbf{x}}^{i}\right\} }_{i=1}^{m}\subset \mathbb{R%
}^{2}$ has the interpolation property for so-called \textit{ridge functions}%
. Ismailov and Pinkus \cite{IP} exploited paths to solve the interpolation
problem on straight lines by ridge functions with fixed directions. In the
special case, when $\mathbf{a}$ and $\mathbf{b}$ are the coordinate vectors
in $\mathbb{R}^{2}$, paths represent \textit{bolts of lightning} (see, e.g.,
\cite{1,6,24}). Note that bolts, first introduced by Diliberto and Straus
\cite{9} under the name of \textit{permissible lines}, played an essential
role in various problems of approximation of multivariate functions by sums
of univariate functions (see, e.g., \cite{9,13,K,23,24}). Note that the name
\textquotedblleft bolt of lightning" is due to Arnold \cite{1}. There is a
useful generalization of closed paths with respect to two directions to
those with respect to finitely many functions. This generalization is
effective in solutions of some representation problems arising in the theory
of linear superpositions (see \cite{Ism}).

In the following, we consider paths with respect to two directions $\mathbf{a%
}$ and $\mathbf{b}$ in $\mathbb{R}^{d}$. A path $\left( \mathbf{p}_{1},%
\mathbf{p}_{2},...,\mathbf{p}_{2n}\right) $ is said to be closed if $\left(
\mathbf{p}_{1},\mathbf{p}_{2},...,\mathbf{p}_{2n},\mathbf{p}_{1}\right) $ is
also a path. The \textit{length} of a path is the number of its points.

We associate each closed path $p=\left( \mathbf{p}_{1},\mathbf{p}_{2},...,%
\mathbf{p}_{2n}\right) $ with the functional
\begin{equation*}
G_{p}(f)=\frac{1}{2n}\sum\limits_{k=1}^{2n}(-1)^{k+1}f(\mathbf{p}_{k}).
\end{equation*}

In the sequel, we will assume that the considered compact set $Q\subset
\mathbb{R}^{d}$ contains a closed path. This assumption is not too
restrictive. Sufficiently many sets in $\mathbb{R}^{d}$ have this property.
For example, any compact set with at least one interior point contains
closed paths. Note that if $Q$ does not contain closed paths, then in almost
all cases we have $E\left( f,\mathcal{N}(\sigma )\right) =0$ for any $f\in
C(Q)$ (see \cite{Ismailov1}). We say \textquotedblleft in almost all cases"
because there is a highly nontrivial example of such $Q$ and continuous $%
f:Q\rightarrow $ $\mathbb{R}$, for which $E\left( f,\mathcal{N}(\sigma
)\right) >0$ (see \cite{Ismailov1}).

We also need the concept of \textit{extremal paths}.

\bigskip

\textbf{Definition 2.2} (see \cite{Is1}). \textit{A finite or infinite path $%
(\mathbf{p}_{1},\mathbf{p}_{2},...)$ is said to be extremal for a function $%
u\in C(Q)$ if $u(\mathbf{p}_{i})=(-1)^{i}\left\Vert u\right\Vert ,i=1,2,...$
or $u(\mathbf{p}_{i})=(-1)^{i+1}\left\Vert u\right\Vert ,$ $i=1,2,...$ .}

\bigskip

The following definition belongs to Schwartz \cite{Sch}.

\bigskip

\textbf{Definition 2.3} (see \cite{Sch}). \textit{A function $\rho \in C(%
\mathbb{R})$ is called mean periodic if the set $span\{\rho (x-\theta ):\
\theta \in \mathbb{R}\}$ is not dense in $C(\mathbb{R})$ in the topology of
uniform convergence on compacta.}

\bigskip

Properties of mean periodic functions were studied in several papers (see,
e.g., \cite{Kah,Koo,Lai,Sch}). It was proven that the condition in
Definition 2.3 is equivalent to each of the following conditions:

a) there exists a non-zero measure $\mu $ of compact support such that

\begin{equation*}
\int \rho (x-y)\mu (y)=0,
\end{equation*}%
for all $x\in \mathbb{R}$;

b) $\rho $ is the limit in $C(\mathbb{R})$ of a sequence of exponential
polynomials $P(x)e^{i\lambda x}$, which are orthogonal to a measure $\mu $
with compact support, that is,

\begin{equation*}
\int P(y)e^{-i\lambda y}\mu (y)=0.
\end{equation*}

For equivalence of the above conditions and for detailed information on mean
periodic functions see Kahane \cite{Kah}.

In our main result (see Theorem 2.1 below), we assume that the considered
function $f$ has a best approximation in the set
\begin{equation*}
\mathcal{R}(\mathbf{a},\mathbf{b})=\left\{ g(\mathbf{a}\cdot \mathbf{x})+h(%
\mathbf{b}\cdot \mathbf{x}):\text{ }g,h\in C(\mathbb{R})\right\} ,
\end{equation*}%
that is, there exists $v_{0}\in \mathcal{R}(\mathbf{a},\mathbf{b})$ such that

\begin{equation*}
\left\Vert f-v_{0}\right\Vert =\inf_{v\in \mathcal{R}(\mathbf{a},\mathbf{b}%
)}\left\Vert f-v\right\Vert .
\end{equation*}%
Some results on existence of a best approximation from $\mathcal{R}(\mathbf{a%
},\mathbf{b)}$ was obtained in our paper \cite{Is3}.

The following lower bound error estimate holds in approximation with
elements from $\mathcal{N}(\sigma )$.

\bigskip

\textbf{Lemma 2.1}. \textit{Assume $\sigma $ is an arbitrary continuous
activation function. Then}
\begin{equation*}
\sup\limits_{p\subset Q}\left\vert G_{p}(f)\right\vert \leq E\left( f,%
\mathcal{N}(\sigma )\right) ,\eqno(2.1)
\end{equation*}%
\textit{for any $f\in C(Q)$. Here the sup is taken over all closed paths.}

\bigskip

\begin{proof} Consider an element of $\mathcal{N}(\sigma )$. This is a sum of
the functions $f_{i}(\mathbf{x})=c_{i}\sigma (\mathbf{w}^{i}\mathbf{\cdot x}%
-\theta _{i}),$ $i=1,...,r.$ Note that for each $c_{i},\theta _{i}\in
\mathbb{R}$, $f_{i}(\mathbf{x})$ is a function of the form $g(\mathbf{w}%
^{i}\cdot \mathbf{x}).$ Since the weight $\mathbf{w}^{i}=$ $\mathbf{a}$ or $%
\mathbf{w}^{i}=$ $\mathbf{b}$, we have $g(\mathbf{w}^{i}\cdot \mathbf{x})=g(%
\mathbf{a}\cdot \mathbf{x)}$ or $g(\mathbf{w}^{i}\cdot \mathbf{x})=g(\mathbf{%
b}\cdot \mathbf{x)}$. Thus, any neural network $\sum_{i=1}^{r}c_{i}\sigma (%
\mathbf{w}^{i}\mathbf{\cdot x}-\theta _{i})$ in $\mathcal{N}(\sigma )$ is an
element of $\mathcal{R}(\mathbf{a},\mathbf{b})$.

Assume $p$ is a closed path in $Q$ and $\Lambda $ is an arbitrary network
from $\mathcal{N}(\sigma )$. Since $\Lambda (\mathbf{x})=g(\mathbf{a}\cdot
\mathbf{x)}+h(\mathbf{b}\cdot \mathbf{x),}$ it is not difficult to verify
that $G_{p}(\Lambda )=0.$ On the other hand, from the definition of $G_{p}$,
it follows that $\left\Vert G_{p}\right\Vert \leq 1$. Thus we obtain that
\begin{equation*}
\left\vert G_{p}(f)\right\vert =\left\vert G_{p}(f-\Lambda )\right\vert \leq
\left\Vert f-\Lambda \right\Vert .
\end{equation*}%
Since the left-hand side and the right-hand side of this inequality do not
depend on $\Lambda $ and $p$, respectively, it follows that
\begin{equation*}
\sup_{p\subset Q}\left\vert G_{p}(f)\right\vert \leq \inf_{\Lambda \in
\mathcal{N}(\sigma )}\left\Vert f-\Lambda \right\Vert =E(f,\mathcal{N}%
(\sigma )).
\end{equation*}
\end{proof}

The following theorem is valid.

\bigskip

\textbf{Theorem 2.1.} \textit{Assume $Q\subset \mathbb{R}^{d}$ is a compact
set and $f\in C(Q).$ Suppose the following conditions hold.}

\textit{1) $f$ has a best approximation in $\mathcal{R}(\mathbf{a},\mathbf{b)%
}$;}

\textit{2) There exists a positive integer $N$ such that any path $(\mathbf{p%
}_{1},...,\mathbf{p}_{n})\subset Q,$ $n>N,\mathit{\ }$or a subpath of it can
be made closed by adding not more than $N$ points of $Q$.}

\textit{Then for any activation function $\sigma $, which is not mean
periodic, the approximation error of the class of single hidden layer
networks $\mathcal{N}(\sigma )$ can be computed by the formula
\begin{equation*}
E\left( f,\mathcal{N}(\sigma )\right) =\sup\limits_{p\subset Q}\left\vert
G_{p}(f)\right\vert ,
\end{equation*}%
where the sup is taken over all closed paths.}

\bigskip

\begin{proof} By assumption, $f$ has a best approximation in $\mathcal{R}(%
\mathbf{a},\mathbf{b)}$. Denote this function by $v_{0}(\mathbf{x})=g_{0}(%
\mathbf{a}\cdot \mathbf{x)}+h_{0}(\mathbf{b}\cdot \mathbf{x)}$. Let us
concentrate on extremal paths for the function $f_{1}=f-v_{0}$. The main
result of \cite{Is1} says that regarding such paths there may be only two
cases.

\textbf{Case 1.} There exists a closed path $p_{0}=\left( \mathbf{p}_{1},...,%
\mathbf{p}_{2n}\right) $ extremal for the function $f_{1}.$

In this case, based on Definition 2.2, we can write that

\begin{equation*}
\left\vert G_{p_{0}}(f)\right\vert =\left\vert G_{p_{0}}(f-v_{0})\right\vert
=\left\Vert f-v_{0}\right\Vert .\eqno(2.2)
\end{equation*}

Since $\sigma $ is not mean periodic, the $span\{\sigma (x-\theta ):\ \theta
\in \mathbb{R}\}$ is dense in $C(\mathbb{R})$ in the topology of uniform
convergence on compacta. It follows that for any $\varepsilon >0$ there
exist natural numbers $m_{1},m_{2}$ and real numbers $c_{ij},\theta _{ij}$, $%
i=1,2$, $j=1,...,m_{i},$ for which

\begin{equation*}
\left\vert g_{0}(t)-\sum_{j=1}^{m_{1}}c_{1j}\sigma (t-\theta
_{1j})\right\vert <\frac{\varepsilon }{2}\eqno(2.3)
\end{equation*}%
and

\begin{equation*}
\left\vert h_{0}(t)-\sum_{j=1}^{m_{2}}c_{2j}\sigma (t-\theta
_{2j})\right\vert <\frac{\varepsilon }{2}\eqno(2.4)
\end{equation*}%
for all $t\in \lbrack a,b]$. Here $[a,b]$ is a sufficiently large interval
which contains both the sets $\{\mathbf{a}\cdot \mathbf{x}:\mathbf{x}\in Q\}$
and $\{\mathbf{b}\cdot \mathbf{x}:\mathbf{x}\in Q\}$.

Taking $t=\mathbf{a}\cdot \mathbf{x}$ in (2.3) and $t=\mathbf{b}\cdot
\mathbf{x}$ in (2.4) we obtain that

\begin{equation*}
\left\vert g_{0}(\mathbf{a}\cdot \mathbf{x)}+h_{0}(\mathbf{b}\cdot \mathbf{x)%
}-\sum_{i=1}^{m}c_{i}\sigma \left( \mathbf{w}^{i}\cdot \mathbf{x}-\theta
_{i}\right) \right\vert <\varepsilon ,\eqno(2.5)
\end{equation*}%
for all $\mathbf{x}\in Q$ and some $c_{i},\theta _{i}\in \mathbb{R}$ and $%
\mathbf{w}^{i}=$ $\mathbf{a}$ or $\mathbf{w}^{i}=$ $\mathbf{b}$. Clearly,

\begin{equation*}
\left\Vert f-\sum_{i=1}^{m}c_{i}\sigma \left( \mathbf{w}^{i}\cdot \mathbf{x}%
-\theta _{i}\right) \right\Vert
\end{equation*}

\begin{equation*}
\leq \left\Vert f-g_{0}-h_{0}\right\Vert +\left\Vert
g_{0}+h_{0}-\sum_{i=1}^{m}c_{i}\sigma \left( \mathbf{w}^{i}\cdot \mathbf{x}%
-\theta _{i}\right) \right\Vert .\eqno(2.6)
\end{equation*}%
It follows from (2.6) that

\begin{equation*}
E\left( f,\mathcal{N}(\sigma )\right) \leq \left\Vert
f-g_{0}-h_{0}\right\Vert +\left\Vert g_{0}+h_{0}-\sum_{i=1}^{m}c_{i}\sigma
\left( \mathbf{w}^{i}\cdot \mathbf{x}-\theta _{i}\right) \right\Vert .\eqno%
(2.7)
\end{equation*}%
The last inequality together with (2.2) and (2.5) yield

\begin{equation*}
E\left( f,\mathcal{N}(\sigma )\right) \leq \left\vert
G_{p_{0}}(f)\right\vert +\varepsilon .
\end{equation*}%
Now since $\varepsilon $ is arbitrarily small, we obtain that
\begin{equation*}
E\left( f,\mathcal{N}(\sigma )\right) \leq \left\vert
G_{p_{0}}(f)\right\vert .
\end{equation*}%
From this and Lemma 2.1 it follows that

\begin{equation*}
E\left( f,\mathcal{N}(\sigma )\right) =\sup\limits_{p\subset Q}\left\vert
G_{p}(f)\right\vert ,
\end{equation*}%
where the $\sup $ is taken over all closed paths.

\textbf{Case 2.} There exists an infinite path extremal for $f_{1}$. Assume
a path $p=(\mathbf{p}_{1},\mathbf{p}_{2},...)$ is infinite and extremal for $%
f_{1}$. Then by the assumption of the theorem, the finite extremal paths $(%
\mathbf{p}_{1},\mathbf{p}_{2},...,\mathbf{p}_{n})\subset p$, $n=N+1,N+2,...$%
, or subpaths of them must be made closed by adding not more than $N$
points. Without loss of generality we may assume that these paths themselves
can be made closed. That is, for each finite extremal path $p_{n}=(\mathbf{p}%
_{1},\mathbf{p}_{2},...,\mathbf{p}_{n})$, $n>N$, there exists a closed path $%
l_{n}=(\mathbf{p}_{1},\mathbf{p}_{2},...,\mathbf{p}_{n},\mathbf{q}_{n+1},...,%
\mathbf{q}_{n+m_{n}})$, where $m_{n}\leq N$. The functional $G_{l_{n}}$
obeys the inequalities
\begin{equation*}
\left\vert G_{l_{n}}(f)\right\vert =\left\vert G_{l_{n}}(f-v_{0})\right\vert
\leq \frac{n\left\Vert f-v_{0}\right\Vert +m_{n}\left\Vert
f-v_{0}\right\Vert }{n+m_{n}}=\left\Vert f-v_{0}\right\Vert \eqno(2.8)
\end{equation*}%
and
\begin{equation*}
\left\vert G_{l_{n}}(f)\right\vert \geq \frac{n\left\Vert f-v_{0}\right\Vert
-m_{n}\left\Vert f-v_{0}\right\Vert }{n+m_{n}}=\frac{n-m_{n}}{n+m_{n}}%
\left\Vert f-v_{0}\right\Vert .\eqno(2.9)
\end{equation*}%
We obtain from (2.8) and (2.9) that
\begin{equation*}
\sup_{l_{n}}\left\vert G_{l_{n}}(f)\right\vert =\left\Vert
f-v_{0}\right\Vert .\eqno(2.10)
\end{equation*}

Using the above sum $\sum_{i=1}^{m}c_{i}\sigma \left( \mathbf{w}^{i}\cdot
\mathbf{x}-\theta _{i}\right) $ and the inequalities (2.5) with (2.7) here,
we obtain from (2.10) that

\begin{equation*}
E\left( f,\mathcal{N}(\sigma )\right) \leq \sup_{l_{n}}\left\vert
G_{l_{n}}(f)\right\vert .\eqno(2.11)
\end{equation*}%
The inequality (2.11) together with (2.1) yield that

\begin{equation*}
E\left( f,\mathcal{N}(\sigma )\right) =\sup\limits_{p\subset Q}\left\vert
G_{p}(f)\right\vert ,
\end{equation*}%
where the $\sup $ is taken over all closed paths. The theorem has been
proved. \end{proof}

\bigskip

\textbf{Corollary 2.1.} \textit{Let $Q\subset \mathbb{R}^{d}$ be a compact
set, $f\in C(Q)$ and the space $\mathcal{R}\left( \mathbf{a}{,}\mathbf{b}%
\right) $ be proximinal in $C(Q)$ (that is, for any $u\in C(Q)$ there exists
a best approximation in $\mathcal{R}\left( \mathbf{a}{,}\mathbf{b}\right) $%
). Let $\sigma $ be any activation function, which is not mean periodic.
Then the approximation error of the class of single hidden layer networks $%
\mathcal{N}(\sigma )$ can be computed by the formula
\begin{equation*}
E\left( f,\mathcal{N}(\sigma )\right) =\sup\limits_{p\subset Q}\left\vert
G_{p}(f)\right\vert ,
\end{equation*}%
where the sup is taken over all closed paths.}

\bigskip

\begin{proof} Since $\mathcal{R}\left( \mathbf{a}{,}\mathbf{b}\right) $ is
proximinal in $C(Q),$ the lengths of \textit{irreducible paths} are
uniformly bounded by some positive integer $N$ (see \cite{Is3}). Note that a
path $\left( \mathbf{q}_{1},...,\mathbf{q}_{m}\right) $ is irreducible if
there is not a path connecting $\mathbf{q}_{1}$ and $\mathbf{q}_{m}$ with
the length less than $m$. Take any path $p=\left( \mathbf{p}_{1},\mathbf{p}%
_{2},...,\mathbf{p}_{n}\right) $ with the length $n>N.$ Since $n>N$, the
path $p$ is not irreducible. Thus we can join the points $\mathbf{p}_{1}$
and $\mathbf{p}_{n}$ by an irreducible path $q=\left( \mathbf{q}_{1},\mathbf{%
q}_{2}...,\mathbf{q}_{m}\right) $, where $\mathbf{q}_{1}=\mathbf{p}_{1}$ and
$\mathbf{q}_{m}=\mathbf{p}_{n}$. Note that by the proximinality assumption, $%
m\leq N$. Then the ordered set $\left( \mathbf{p}_{1},\mathbf{p}_{2},...,%
\mathbf{p}_{n},\mathbf{q}_{m-1},...\mathbf{q}_{2}\right) $ (or some subset $%
\left( \mathbf{p}_{i},\mathbf{p}_{i+1},...,\mathbf{p}_{k},\mathbf{q}_{j},...%
\mathbf{q}_{s}\right) $ of it) is a closed path, where the number of added
points is less than $N$. We see that all the conditions of Theorem 2.1 are
satisfied; hence the assertion of Corollary 2.1 is valid. \end{proof}

\bigskip

Many activation functions exploited in neural network theory and
applications are not mean periodic. For example, this is true for a number
of popular activation functions (such as sigmoid, hyperbolic tangent,
Gaussian, etc). The following corollary specifies one class of such
functions.

\bigskip

\textbf{Corollary 2.2.} \textit{Assume all the conditions of Theorem 2.1
hold. Let $\sigma \in C(\mathbb{R})\cap L_{p}(\mathbb{R)}$, where $1\leq
p<\infty $, or $\sigma $ be a continuous, bounded, nonconstant function,
which has a limit at infinity (or minus infinity). Then the approximation
error of the class of single hidden layer networks $\mathcal{N}(\sigma )$
can be computed by the formula
\begin{equation*}
E\left( f,\mathcal{N}(\sigma )\right) =\sup\limits_{p\subset Q}\left\vert
G_{p}(f)\right\vert ,
\end{equation*}%
where the sup is taken over all closed paths.}

\bigskip

The proof can be easily obtained from Theorem 2.1 and the following result
of Schwartz \cite{Sch}: Any continuous and $p$-th degree Lebesgue integrable
univariate function or continuous, bounded, nonconstant function having a
limit at infinity (or minus infinity) is not mean periodic (see also \cite%
{A20}).

\bigskip

As an example we show that the $\sup \left\vert G_{p}(f)\right\vert $ in
Theorem 2.1 can be easily computed for some class of functions $f$. For the
sake of simplicity let the space dimension $d=2$. Assume we are given
linearly independent vectors $\mathbf{a}=(a_{1},a_{2})$ and $\mathbf{b}%
=(b_{1},b_{2})$, and the domain
\begin{equation*}
Q=\left\{ \mathbf{x}\in \mathbb{R}^{2}:c_{1}\leq \mathbf{a}\cdot \mathbf{x}%
\leq d_{1},\ \ c_{2}\leq \mathbf{b}\cdot \mathbf{x}\leq d_{2}\right\} ,
\end{equation*}%
where $c_{1}<d_{1}$ and $c_{2}<d_{2}$.

Consider the class $M(Q)$ of continuous functions $f$ on $Q$, which have the
continuous partial derivatives $\frac{\partial ^{2}f}{\partial x_{1}^{2}},%
\frac{\partial ^{2}f}{\partial x_{1}\partial x_{2}},\frac{\partial ^{2}f}{%
\partial x_{2}^{2}}$, and for any $\mathbf{x}=(x_{1},x_{2})\in Q$,

\begin{equation*}
\frac{\partial ^{2}f}{\partial x_{1}\partial x_{2}}\left(
a_{1}b_{2}+a_{2}b_{1}\right) -\frac{\partial ^{2}f}{\partial x_{1}^{2}}%
a_{2}b_{2}-\frac{\partial ^{2}f}{\partial x_{2}^{2}}a_{1}b_{1}\geq 0.\eqno%
(2.12)
\end{equation*}

Using Theorem 2.1 we want to compute the error in approximating $f\in M(Q)$
by elements of the set

\begin{equation*}
\mathcal{N}(\sigma )=span\left\{ \sigma (\mathbf{w\cdot x}-\theta ):\theta
\in \mathbb{R},\text{ }\mathbf{w}=\mathbf{a}\text{ or }\mathbf{w}=\mathbf{b}%
\right\} .
\end{equation*}%
Here $\sigma $ is any non-mean periodic activation function (for example,
any continuous nonconstant function having limits at plus and minus
infinities). Note that all the assumptions of Theorem 2.1 hold, moreover the
set $\mathcal{R}(\mathbf{a},\mathbf{b)}$ is proximinal in $C(Q)$ (see \cite%
{Is3}).

Consider the following linear transformation
\begin{equation*}
y_{1}=a_{1}x_{1}+a_{2}x_{2},\ \ y_{2}=b_{1}x_{1}+b_{2}x_{2}.\eqno(2.13)
\end{equation*}%
Let
\begin{equation*}
K=[c_{1},d_{1}]\times \lbrack c_{2},d_{2}].
\end{equation*}%
Since the vectors $(a_{1},a_{2})$ and $(b_{1},b_{2})$ are linearly
independent, for any $(y_{1},y_{2})\in K$ there exists only one solution $%
(x_{1},x_{2})\in Q$ of the system (2.13). This solution is given by the
formulas

\begin{equation*}
x_{1}=\frac{y_{1}b_{2}-y_{2}a_{2}}{a_{1}b_{2}-a_{2}b_{1}},\qquad \ x_{2}=%
\frac{y_{2}a_{1}-y_{1}b_{1}}{a_{1}b_{2}-a_{2}b_{1}}.\eqno(2.14)
\end{equation*}

The linear transformation (2.14) transforms the function $f(x_{1},x_{2})$ to
the function $g(y_{1},y_{2})$. Besides, this transformation maps paths with
respect to the directions $(a_{1},a_{2})$ and $(b_{1},b_{2})$ to paths with
respect to the coordinate directions $(1,0)$ and $(0,1)$. As we have already
known the latter type of paths are called lightning bolts (see Definition
2.1 and the subsequent discussions). Hence,%
\begin{equation*}
\sup\limits_{p\subset Q}\left\vert G_{p}(f)\right\vert
=\sup\limits_{q\subset K}\left\vert G_{q}(g)\right\vert ,\eqno(2.15)
\end{equation*}%
where the $\sup $ in the left hand side of (2.15) is taken over closed paths
with respect to the directions $(a_{1},a_{2})$ and $(b_{1},b_{2})$, while
the $\sup $ in the right hand side of (2.15) is taken over closed bolts.

Note that
\begin{equation*}
\frac{\partial ^{2}g}{\partial y_{1}\partial y_{2}}\geq 0,\eqno(2.16)
\end{equation*}%
for any $(y_{1},y_{2})\in K$, which easily follows from (2.12).

The $\sup $ in the right hand side of (2.15) can be computed by applying
theorems of Ofman \cite{Ofm}, and Rivlin and Sibner \cite{Riv}. By Ofman's
theorem

\begin{equation*}
\sup\limits_{q\subset K}\left\vert G_{q}(g)\right\vert
=\inf_{g_{1}+g_{2}}\left\Vert
g(y_{1},y_{2})-g_{1}(y_{1})-g_{2}(y_{2})\right\Vert _{C(K)}.\eqno(2.17)
\end{equation*}%
By a result of Rivlin and Sibner (see \cite{Riv}), Eq. (2.16) yields that

\begin{equation*}
\inf_{g_{1}+g_{2}}\left\Vert
g(y_{1},y_{2})-g_{1}(y_{1})-g_{2}(y_{2})\right\Vert _{C(K)}=\frac{1}{4}%
\iint\nolimits_{K}\frac{\partial ^{2}g}{\partial y_{1}\partial y_{2}}%
dy_{1}dy_{2}.\eqno(2.18)
\end{equation*}%
It follows from Corollary 2.1 and equations (2.15), (2.17) and (2.18) that

\begin{equation*}
E\left( f,\mathcal{N}(\sigma )\right) =\frac{1}{4}\iint\nolimits_{K}\frac{%
\partial ^{2}g}{\partial y_{1}\partial y_{2}}dy_{1}dy_{2}.\eqno(2.19)
\end{equation*}%
The above integral can be computed easily, using values of $g$ at the
vertices of $K$.

\bigskip

\textbf{Example.} Let $\mathbf{a}=(a_{1},a_{2})$ and $\mathbf{b}%
=(b_{1},b_{2})$ be the coordinate vectors $(1,0)$ and $(0,1),$ respectively.
Assume $Q$ is the unit square $[0,1]^{2}$ and $\sigma $ is a sigmoidal
function. Assume we are given the function  $f(x_{1},x_{2})=x_{1}x_{2}.$
Note that (2.12) holds, hence $f\in M(Q)$. The approximating set of networks
$\mathcal{N}(\sigma )$ has members of the form

\begin{equation*}
\sum_{i=1}^{n_{1}}c_{i}\sigma (x_{1}-\theta
_{i})+\sum_{j=1}^{n_{2}}d_{j}\sigma (x_{2}-\lambda _{j}),
\end{equation*}%
where $c_{i},d_{j},\theta _{i},\lambda _{j}$ are arbitrary real numbers and $%
n_{1},n_{2}$ are positive integers. Since linear transformation (2.13) does
not change the coordinates in our case, we have $g=f$ and $K=Q.$ Thus, by
formula (2.19),

\begin{equation*}
E\left( f,\mathcal{N}(\sigma )\right) =\frac{1}{4}\iint\nolimits_{Q}\frac{%
\partial ^{2}f}{\partial x_{1}\partial x_{2}}dx_{1}dx_{2}=\frac{1}{4}.
\end{equation*}

\bigskip

\textbf{Remark.} The question on computing the approximation error of neural
nets with more than two fixed weights is fair, but its solution seems to be
beyond the scope of the methods discussed herein. A path with respect to two
directions $\mathbf{a}$ and $\mathbf{b}$ is constructed as an ordered set of
points $\left( \mathbf{p}_{1},\mathbf{p}_{2},...,\mathbf{p}_{n}\right) $ in $%
\mathbb{R}^{d}$ with edges $\mathbf{p}_{i}\mathbf{p}_{i+1}$ in alternating
hyperplanes so that the first, third, fifth and so on hyperplanes (also the
second, fourth, sixth and so on hyperplanes) are parallel. If not
differentiate between parallel hyperplanes, the path $\left( \mathbf{p}_{1},%
\mathbf{p}_{2},...,\mathbf{p}_{n}\right) $ can be considered as a trace of
some point traveling in two alternating hyperplanes. In this case, the path
functional
\begin{equation*}
F(f)=\frac{1}{n}\sum\limits_{i=1}^{n}(-1)^{i+1}f(\mathbf{p}_{i}),
\end{equation*}%
has some important properties, which lead to a geometric criterion for a
best approximation from $\mathcal{R}(\mathbf{a},\mathbf{b)}$ (see \cite{Is1}%
). Note that our Theorem 2.1 is mainly based on this criterion. The problem
becomes complicated when the number of directions is more than two. The
simple generalization of paths demands a point traveling in three or more
alternating hyperplanes. But in this case the appropriate generalization of
the above functional $F$ looses its original useful properties. Some
difficulties with a generalization of paths and path functionals were
delineated in \cite{Ism} and \cite{Is1}.


\bigskip

\begin{thebibliography}{99}
\bibitem{1} V.I. Arnold, On functions of three variables, \textit{Dokl.
Akad. Nauk SSSR} \textbf{114} (1957), 679-681; English transl, \textit{Amer.
Math. Soc. Transl.} \textbf{28} (1963), 51-54.

\bibitem{Barron} A.R. Barron, Universal approximation bounds for
superposition of a sigmoidal function, \textit{IEEE Trans. Information Theory%
} \textbf{39 }(1993), 930--945.

\bibitem{3} D. Braess and A. Pinkus, Interpolation by ridge functions,
\textit{J.Approx. Theory} \textbf{73} (1993), 218-236.

\bibitem{6} E.J. Cand\'{e}s, Ridgelets: estimating with ridge functions,
\textit{Ann. Statist.} \textbf{31} (2003), 1561-1599.

\bibitem{A2} T. Chen and H. Chen, Approximation of continuous functionals by
neural networks with application to dynamic systems, \textit{IEEE Trans.
Neural Networks} \textbf{4} (1993), 910-918.

\bibitem{A3} C.K. Chui and X. Li, Approximation by ridge functions and
neural networks with one hidden layer, \textit{J. Approx. Theory} \textbf{70}
(1992), 131-141.

\bibitem{Cos1} D. Costarelli and R. Spigler, Approximation results for
neural network operators activated by sigmoidal functions, \textit{Neural
Networks} \textbf{44} (2013), 101-106.

\bibitem{Cos2} D. Costarelli and G. Vinti, Saturation classes for
max-product neural network operators activated by sigmoidal functions,
\textit{Results Math.} \textbf{72} (2017), no.3, 1555--1569.

\bibitem{A4} G. Cybenko, Approximation by superpositions of a sigmoidal
function, \textit{Math. Control, Signals, and Systems} \textbf{2} (1989),
303-314.

\bibitem{9} S.P. Diliberto and E.G. Straus, On the approximation of a
function of several variables by the sum of functions of fewer variables,
\textit{Pacific J.Math.} \textbf{1} (1951), 195-210.

\bibitem{13} M.v. Golitschek and W.A. Light, Approximation by solutions of
the planar wave equation, \textit{Siam J.Numer. Anal.} \textbf{29} (1992),
816-830.

\bibitem{GI2} N.J. Guliyev and V.E. Ismailov, On the approximation by single
hidden layer feedforward neural networks with fixed weights, \textit{Neural
Networks} \textbf{98} (2018), 296--304.

\bibitem{GI3} N.J. Guliyev and V.E. Ismailov, Approximation capability of
two hidden layer feedforward neural networks with fixed weights, \textit{%
Neurocomputing} \textbf{316} (2018), 262--269.

\bibitem{Hahm} N. Hahm and B.I. Hong, Extension of localized approximation
by neural networks, \textit{Bull. Austral. Math. Soc.} \textbf{59} (1999),
121--131.

\bibitem{A7} K. Hornik, Approximation capabilities of multilayer feedforward
networks, \textit{Neural Networks} \textbf{4} (1991), 251-257.

\bibitem{Il} A. Iliev, N. Kyurkchiev and S. Markov, Approximation of the cut
function by Stannard and Richard sigmoid functions, \textit{Int. J. Pure
Appl. Math.} \textbf{109} (2016), no. 1, 119-128.

\bibitem{Ism} V.E. Ismailov, \textit{Ridge Functions and Applications in
Neural Networks}, Mathematical Surveys and Monographs, 263. American
Mathematical Society, 2021.

\bibitem{Is1} V.E. Ismailov, A note on the equioscillation theorem for best
ridge function approximation, \textit{Expo. Math.} \textbf{35} (2017), no.
3, 343-349.

\bibitem{Ismailov1} V.E. Ismailov, Approximation by neural networks with
weights varying on a finite set of directions. \textit{J. Math. Anal. Appl.}
\textbf{389} (2012), no. 1, 72--83.

\bibitem{Is3} V.E. Ismailov, On the proximinality of ridge functions,
\textit{Sarajevo J. Math.} \textbf{5 (17)} (2009), no. 1, 109-118.

\bibitem{IP} V.E. Ismailov and A. Pinkus, Interpolation on lines by ridge
functions, \textit{J. Approx. Theory} \textbf{175} (2013), 91-113.

\bibitem{IS} V.E. Ismailov and E. Savas, Measure theoretic results for
approximation by neural networks with limited weights, \textit{Numer. Funct.
Anal. Optim.} \textbf{38} (2017), no. 7, 819-830.

\bibitem{A10} Y. Ito, Approximation of continuous functions on $\mathbb{R}%
^{d}$ by linear combinations of shifted rotations of a sigmoid function with
and without scaling, \textit{Neural Networks} \textbf{5} (1992), 105-115.

\bibitem{Kah} J.P. Kahane, \textit{Lectures on Mean Periodic Functions},
Tata Institute of Fundamental Research, Bombay 1959.

\bibitem{K} S.Ya. Khavinson, \textit{Best approximation by linear
superpositions (approximate nomography),} Translated from the Russian
manuscript by D. Khavinson. Translations of Mathematical Monographs, 159.
American Mathematical Society, Providence, RI, 1997, 175 pp.

\bibitem{Koo} P. Koosis, On functions which are mean periodic on a half
line, \textit{Comm. Pure Appl. Math.} \textbf{10} (1957), 133-149.

\bibitem{Ky} N. Kyurkchiev and S. Markov, \textit{Sigmoid functions: some
approximation and modelling aspects}, Lambert Academic Publishing,
Saarbrucken, 2015.

\bibitem{Lai} P.G. Laird, Some properties of mean periodic functions,
\textit{J. Austral. Math. Soc.} \textbf{14} (1972), 424-432.

\bibitem{A13} M. Leshno, V. Ya. Lin, A. Pinkus and S. Schocken, Multilayer
feedforward networks with a nonpolynomial activation function can
approximate any function, \textit{Neural Networks} \textbf{6} (1993),
861-867.

\bibitem{Lin1} S. Lin, X. Guo, F. Cao and Z. Xu, Approximation by neural
networks with scattered data, \textit{Appl. Math. Comput.} \textbf{224}
(2013), 29-35.

\bibitem{Lin2} S. Lin, J. Zeng, L. Xu and Z. Xu, Jackson-type inequalities
for spherical neural networks with doubling weights, \textit{Neural Networks}
\textbf{63} (2015), 57-65.

\bibitem{Maiorov} V. Maiorov and R.S. Meir, Approximation bounds for smooth
functions in $C(\mathbb{R}^{d})$ by neural and mixture networks, \textit{%
IEEE Trans. Neural Networks} \textbf{9} (1998), 969-978.

\bibitem{MP} V. Maiorov and A. Pinkus, Lower bounds for approximation by MLP
neural networks, \textit{Neurocomputing} \textbf{25} (1999), 81-91.

\bibitem{23} D.E. Marshall and A.G. O'Farrell. Uniform approximation by real
functions, \textit{Fund. Math.} \textbf{104} (1979), 203-211.

\bibitem{24} D.E. Marshall and A.G. O'Farrell, Approximation by a sum of two
algebras. The lightning bolt principle, \textit{J. Funct. Anal.} \textbf{52}
(1983), 353-368.

\bibitem{Mhaskar} H.N. Mhaskar and C.A. Micchelli, Degree of approximation
by neural networks with a single hidden layer, \textit{Adv. Appl. Math.}
\textbf{16} (1995), 151-183.

\bibitem{Ofm} Ju.P. Ofman, On the best approximation of functions of two
variables by functions of the form $\varphi (x)+\psi (y)$, (Russian) \textit{%
Izv. Akad. Nauk SSSR Ser. Mat.} \textbf{25} (1961), 239-252.

\bibitem{A20} A. Pinkus, Approximation theory of the MLP model in neural
networks, \textit{Acta Numerica} \textbf{8} (1999), 143-195.

\bibitem{Riv} T.J. Rivlin and R.J. Sibner, The degree of approximation of
certain functions of two variables by a sum of functions of one variable,
\textit{Amer. Math. Monthly} \textbf{72} (1965), 1101-1103.

\bibitem{Sch} L. Schwartz, Theorie generale des fonctions
moyenne-periodiques, Ann. Math. \textbf{48} (1947), 857-928.

\bibitem{A25} M. Stinchcombe and H. White, Approximating and learning
unknown mappings using multilayer feedforward networks with bounded weights,
in \textit{Proceedings of the IEEE 1990 International Joint Conference on
Neural Networks}, 1990, Vol. 3, IEEE, New York, 7-16.
\end{thebibliography}
\end{document}